\pdfoutput=1
\documentclass[11pt]{article}

\PassOptionsToPackage{table}{xcolor}
\usepackage[preprint]{acl}

\usepackage{times}
\usepackage{latexsym}
\usepackage[T1]{fontenc}
\usepackage[utf8]{inputenc}
\usepackage{microtype}
\usepackage{inconsolata}
\usepackage{graphicx}

\usepackage{amsmath}
\usepackage{amssymb}
\usepackage{mathtools}
\usepackage{booktabs}
\usepackage{multirow}
\usepackage{siunitx}
\usepackage{colortbl}
\usepackage{algorithm}
\usepackage{algorithmic}
\usepackage{varwidth}
\usepackage[most]{tcolorbox}
\usepackage{helvet}
\usepackage[capitalize,noabbrev]{cleveref}

\usepackage{amsmath,amsfonts,bm}

\def\eqref#1{equation~\ref{#1}}

\def\1{\bm{1}}

\def\vx{{\bm{x}}}
\def\vy{{\bm{y}}}
\def\vz{{\bm{z}}}

\DeclareMathAlphabet{\mathsfit}{\encodingdefault}{\sfdefault}{m}{sl}
\SetMathAlphabet{\mathsfit}{bold}{\encodingdefault}{\sfdefault}{bx}{n}

\def\gZ{{\mathcal{Z}}}

\crefformat{section}{\S#2#1#3}
\crefformat{subsection}{\S#2#1#3}
\crefformat{subsubsection}{\S#2#1#3}
\crefrangeformat{section}{\S\S#3#1#4 to~#5#2#6}
\crefmultiformat{section}{\S\S#2#1#3}{ and~#2#1#3}{, #2#1#3}{ and~#2#1#3}
\Crefformat{figure}{#2Figure~#1#3}
\Crefmultiformat{figure}{Figs.~#2#1#3}{ and~#2#1#3}{, #2#1#3}{ and~#2#1#3}
\Crefformat{table}{#2Table~#1#3}
\Crefmultiformat{table}{Tabs.~#2#1#3}{ and~#2#1#3}{, #2#1#3}{ and~#2#1#3}
\Crefformat{appendix}{Appx.~\S#2#1#3}
\crefformat{algorithm}{Alg.~#2#1#3}
\Crefformat{equation}{Eq.~(#2#1#3)}
\Crefmultiformat{equation}{Eqs.~(#2#1#3)}{ and~(#2#1#3)}{, (#2#1#3)}{ and~(#2#1#3)}

\newcommand{\ourMethod}{DISCO}

\definecolor{avgcol}{gray}{0.95}
\definecolor{grayrowcolor}{gray}{0.90}
\newcommand{\modelrow}[1]{\multicolumn{11}{c}{\cellcolor{grayrowcolor}\textbf{\textsc{#1}}}}

\newtcolorbox{casebox}[1]{%
    enhanced, breakable,
    colback=gray!4, colframe=black!55,
    boxrule=0.4pt, arc=2pt,
    left=8pt, right=8pt, top=5pt, bottom=5pt,
    fonttitle=\bfseries\small, fontupper=\small,
    coltitle=black, colbacktitle=gray!18,
    title={#1},
    attach boxed title to top left={xshift=6pt, yshift=-7pt},
    boxed title style={colframe=black!55, boxrule=0.3pt, arc=1.5pt},
    varwidth boxed title*=-40pt%
}

\title{\ourMethod{}: Distributed Long Context Scaling with\\Grounding-Reasoning Disaggregation}

\author{
  \textbf{Guanzheng Chen\textsuperscript{1,2}}\thanks{Work done during an internship at Adobe Research.},
  \textbf{Viet Dac Lai\textsuperscript{2}},
  \textbf{Subhojyoti Mukherjee\textsuperscript{2}},
  \textbf{Branislav Kveton\textsuperscript{2}},
\\
  \textbf{Seunghyun Yoon\textsuperscript{2}},
  \textbf{Franck Dernoncourt\textsuperscript{2}},
  \textbf{Qizhe Xie\textsuperscript{1}}\thanks{Corresponding authors.},
  \textbf{Trung Bui\textsuperscript{2}}\footnotemark[2]
\\
\\
  \textsuperscript{1}National University of Singapore,
  \textsuperscript{2}Adobe Research
\\
  \small{
    \textbf{Correspondence:} \href{mailto:qizhex@nus.edu.sg}{qizhex@nus.edu.sg}, \href{mailto:bui@adobe.com}{bui@adobe.com}
  }
}

\begin{document}
\maketitle

\begin{abstract}
While Large Language Models (LLMs) advertise million-token context windows, reasoning quality often collapses as inputs grow---a phenomenon termed \emph{context rot}. This failure stems from a structural entanglement in monolithic architectures, where the massive search burden of \textit{contextual grounding} exhausts the representational capacity needed for complex \textit{reasoning}. To resolve this, we propose \textbf{Grounding-Reasoning Disaggregation} via \textbf{DIS}tributed long \textbf{CO}ntext scaling (\textbf{\ourMethod{}}). Inspired by distributed computing frameworks like Apache Spark, \ourMethod{} partitions long context across a fleet of \emph{Worker} LLMs dedicated exclusively to parallel, localized grounding. A central \emph{Driver} LLM, trained via Reinforcement Learning (GRPO) to optimize planning, orchestrates execution by dynamically mapping queries into atomic extraction tasks and reducing the gathered evidence to synthesize a final answer. By isolating reasoning from raw context noise, \ourMethod{} effectively eliminates context rot. On RULER-QA (1M tokens), it maintains 78.4\% accuracy where standard baselines collapse. Furthermore, it outperforms full-context models by up to 9.8 points on LongBench v2 and matches frontier models like Gemini-3-Pro-Preview while reducing inference costs by over 80\%, establishing a highly efficient paradigm for robust long-context inference.
\end{abstract}

\section{Introduction}

Recent Large Language Models (LLMs) increasingly advertise context windows spanning millions of tokens~\citep{openai2023gpt4, huang2025gemini, guo2025deepseek}. Yet, even well within nominal limits, reasoning quality frequently collapses as input grows---a phenomenon termed \emph{context rot}~\citep{hsieh2024ruler, hong2025context}. Consequently, the \textit{effective} context window of LLMs remains severely limited~\citep{modarressi2025nolima}, hampering their reliability in real-world applications.

We argue that context rot stems from a fundamental structural asymmetry between two distinct cognitive phases: \textit{contextual grounding} (identifying relevant facts) and \textit{reasoning} (synthesizing those facts into an answer). As input length increases, grounding becomes a massive search problem that scales proportionally with the text, while the core reasoning task typically remains constant in complexity regardless of the haystack's size. In monolithic Transformer architectures~\citep{vaswani2017attention}, however, these two processes are irreversibly entangled. Attention heads forced to locate scattered needles across millions of tokens exhaust their representational bandwidth, leaving insufficient residual capacity for multi-hop reasoning.

Existing approaches struggle to resolve this entanglement. Retrieval pipelines and agentic memory systems~\citep{chhikara2025mem0,zhou2025mem1,li2025memos} separate retrieval from reasoning, but delegate grounding to shallow mechanisms (e.g., embeddings, BM25) that frequently miss implicitly relevant evidence. Conversely, sequential multi-agent workflows~\citep{zhang2024chain, qian2024long, zhao2024longagent, zhou2024llm} apply full LLM-grade understanding to each chunk, but process them serially---incurring linear latency and fragile cross-step state management that degrades over multiple hops. Ultimately, these paradigms fail to simultaneously achieve deep, parallelizable grounding that is structurally isolated from reasoning.\footnote{Alternative approaches like memory-augmented architectures~\citep{behrouz2024titans} internalize long-range retention but currently demand prohibitive pre-training from scratch, leaving generalization underexplored.}

To scale robustly toward millions of tokens, we propose a structural paradigm shift: \textit{Grounding-Reasoning Disaggregation}. The core principle is to decouple these two phases entirely, ensuring that complex reasoning is conducted exclusively over a highly refined subset of facts extracted during an independent, parallelized grounding stage. By isolating the heavy lifting of information retrieval from the cognitive load of synthesis, we protect the model's reasoning capabilities from being overwhelmed by scale.
To realize this disaggregation, we introduce \textbf{DIS}tributed long \textbf{CO}ntext scaling (\textbf{\ourMethod{}}), an architecture directly inspired by distributed computing frameworks like Apache Spark~\citep{zaharia2010spark,zaharia2016apache}. Conceptually, an LLM's limited effective context window is analogous to the operational memory (RAM) of a single compute node. When data exceeds this capacity, systems like Spark do not build larger nodes; instead, they schedule multiple nodes to process data shards in parallel. Adopting this metaphor, \ourMethod{} partitions the massive context into shards distributed across a fleet of logical \emph{Worker} LLMs. Each Worker persistently holds its own shard to conduct local, high-fidelity grounding. These Workers are orchestrated by a central \emph{Driver} LLM, which functions as a high-level planner and synthesizer.

Inspired by directed acyclic graph (DAG) execution models, the Driver decomposes complex user queries into parallelizable atomic actions. Rather than relying on rigid linear processing, the system maps information-intensive extraction tasks across the Workers to gather facts independently from their local shards, and subsequently reduces these aggregated insights through a global reasoning pass by the Driver to synthesize the final answer.

\ourMethod{} operates dynamically, iteratively re-planning the DAG until sufficient evidence is gathered. This separation enables pure inference-time scaling: as context grows, we scale Map actions; as reasoning complexity increases, we scale Wide-dependency actions. Furthermore, \ourMethod{} provides a native environment for Reinforcement Learning (RL). We employ Group Relative Policy Optimization (GRPO) \citep{shao2024deepseekmath} to train the Driver LLM, incentivizing optimal execution plans and precise reasoning paths.  Experimental results demonstrate that \ourMethod{} effectively eliminates context rot: on RULER-QA (1M tokens), \ourMethod{} with Qwen3-8B~\citep{qwen3technicalreport} maintains 78.4\% accuracy while standard retrieval baselines collapse to 10.9\%. 
On reasoning-intensive benchmarks, it outperforms Full Long Context baselines by up to 9.8 points (Qwen3-14B) on LongBench v2 (Long subset), with RL tuning further amplifying gains. 
Crucially, \ourMethod{} enables economic scaling: it matches the performance of frontier models like Gemini-3-Pro-Preview while reducing inference costs by over 80\%, establishing a new paradigm for efficient, high-fidelity long-context inference.

\section{Method}
\label{sec:method}

We introduce \ourMethod{}, a distributed inference framework that reframes long-context processing as a data-parallel computing problem. Departing from the monolithic paradigm—where a single transient attention mechanism manages millions of tokens—we treat the context as persistent, distributed data. In this section, we derive the theoretical motivation for decoupling contextual grounding from reasoning (\Cref{subsec:method_theory}), propose a distributed architecture to approximate the grounding objective (\Cref{subsec:method_arch}), formalize the inference process as a dynamic Directed Acyclic Graph (DAG) (\Cref{subsec:dag}), and describe the iterative orchestration between Driver and Workers (\Cref{subsec:execution}). Finally, we detail the optimization of the Driver via Reinforcement Learning (\Cref{subsec:rl}).

\subsection{The Grounding-Reasoning Interference}
\label{subsec:method_theory}

Let a long-context task be defined by a massive context sequence $\bm{x} = (x_1, \dots, x_L)$ of length $L$, a query $q$, and a target response $\bm{y}$. Standard LLMs with parameters $\theta$ model the conditional probability $P_\theta(\bm{y} \mid \bm{x}, q)$ monolithically. Intuitively, as $L$ increases, the relevant information for $q$ becomes sparsely distributed across $\bm{x}$. Consequently, the cognitive process implicitly bifurcates into two distinct stages: \textit{contextual grounding} (identifying relevant evidence) and \textit{reasoning} (synthesizing the answer from that evidence).

We formalize this by introducing a latent variable $\bm{z}$, representing the set of discrete evidence artifacts (facts, snippets, or logical predicates) hidden within $\bm{x}$. The generation probability factorizes as:
\begin{equation}
\label{eq:factorization}
      P_\theta(\bm{y} \mid \bm{x}, q) = \sum_{\bm{z}} \underbrace{P_\theta(\bm{y} \mid \bm{z}, q)}_{\text{Reasoning}} \cdot \underbrace{P_\theta(\bm{z} \mid \bm{x}, q)}_{\text{Grounding}}.\footnotemark
\end{equation}

In monolithic architectures, a single set of parameters $\theta$ within a single attention window attempts to approximate both terms simultaneously. This coupling creates a critical bottleneck: as $L$ grows, the complexity of the grounding term $P_\Theta(\bm{z} \!\mid\! \bm{x}, q)$ degrades rapidly due to architectural limitations (e.g., context rot and the ``lost-in-the-middle'' phenomenon), where attention mechanisms fail to distinguish signal from noise. Consequently, the task fails not due to a lack of reasoning capability, but due to a collapse in the evidences grounding.

\subsection{\ourMethod{}: A Distributed Long Context Scaling Paradigm}
\label{subsec:method_arch}

Contextual grounding over massive sequences becomes the primary bottleneck in long-context scaling. To circumvent the limitations of directly modeling the full sequence, we introduce \textbf{Distributed Context Scaling (\ourMethod{})}, a paradigm that structurally decouples contextual grounding via context partitioning. We split the massive context $\vx$ into $K$ disjoint shards $\{\vx^{(k)}\}_{k=1}^K$, ensuring that each shard falls comfortably within the high-fidelity window of a standard LLM. This partitioning allows us to reformulate the grounding term in~\Cref{eq:factorization} as a distributed product:
\begin{equation}
\label{eq:decoupling}
\begin{aligned}
    P_{\theta}(\vz \mid \vx, q) \approx{} & \underbrace{P_{\theta}\left(\vz \;|\; \gZ, q\right)}_{\text{Global Synthesis}} \\
    & \cdot \prod_{k=1}^K \underbrace{P_{\theta}\left(\vz^{(k)} \mid \vx^{(k)}, q\right)}_{\text{Local Grounding}},
\end{aligned}
\end{equation}
where $\gZ = \bigcup_{k=1}^K \vz^{(k)}$ denotes the union of discrete evidence artifacts extracted independently from each shard. Theoretically, this transformation replaces the computationally intractable $O(L^2)$ global extraction with $K$ parallel $O(|\vx^{(k)}|^2)$ local extractions, making inference linearly scalable with respect to context length.

To operationalize this decoupled factorization, we map the components of~\Cref{eq:decoupling} to two distinct system roles: \textbf{Worker} (local grounding) and \textbf{Driver} (global reasoning). While Eq.~\ref{eq:decoupling} enables parallelism, deploying massive LLMs to handle every local grounding task would remain computationally prohibitive. To address this, we adopt a \textbf{Hierarchical Architecture} that pairs a powerful, reasoning-dense model (parameterized by $\theta_{\mathcal{D}}$) for the Driver with a fleet of lightweight, efficient models (parameterized by $\theta_{\mathcal{W}}$) for the Workers. 

Formally, each Worker is tasked to scan its assigned shard $\vx^{(k)}$ to extract local evidence, a process restricted strictly to the local scope:
\begin{equation}
\label{eq:worker_role}
\vz^{(k)} \sim P_{\theta_{\mathcal{W}}}\left( \cdot \mid \vx^{(k)}, q \right).
\end{equation}
Following this extraction, the Driver $\mathcal{D}$ acts as the central reasoning unit. Crucially, the Driver does not access the raw context $\vx$. Instead, it operates on the aggregated outputs from the Workers to first synthesize a global view $\vz$, and subsequently generate the final answer $\vy$:
\begin{equation}
\label{eq:driver_role}
\begin{aligned}
    \text{Synthesis:} \quad & \vz \sim P_{\theta_{\mathcal{D}}}\left(\cdot \;\bigg|\; \bigcup_{k=1}^K \vz^{(k)}, q\right), \\
    \text{Reasoning:} \quad & \vy \sim P_{\theta_{\mathcal{D}}}\left( \cdot \mid \vz, q \right).
\end{aligned}
\end{equation}
By replacing the noise-heavy raw context $\vx$ with high-density evidence $\vz$ in the conditioning term, the Driver restores the fidelity of the attention mechanism, enabling complex reasoning without the interference typically caused by massive input lengths.

\begin{figure*}[t]
    \centering
    \includegraphics[width=0.95\linewidth]{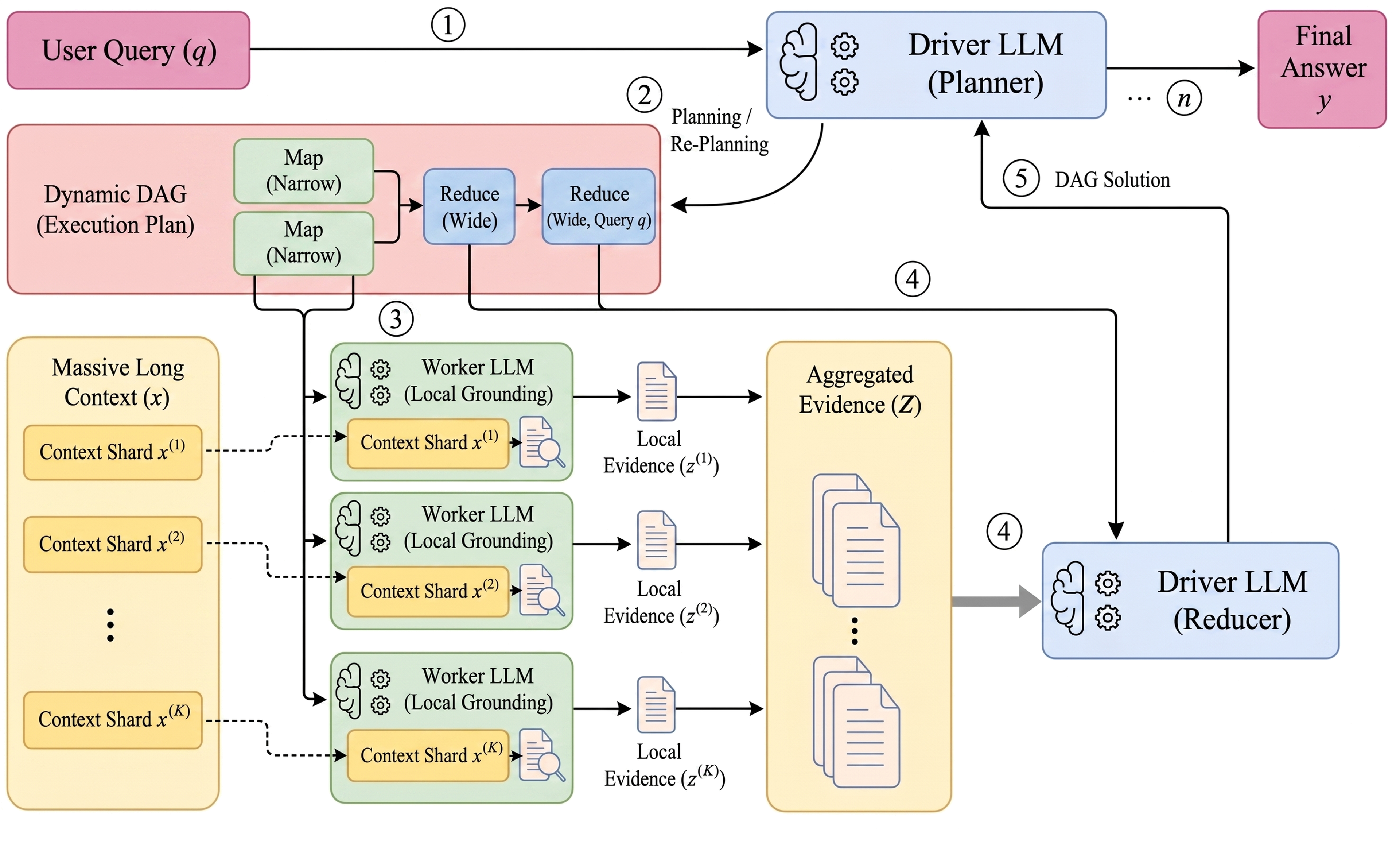}
    \caption{The execution pipeline of~\ourMethod{}.  A Driver LLM orchestrates the inference by dynamically generating a DAG of actions: Map (Narrow) tasks are offloaded to parallel Worker LLMs for evidence extraction, while Reduce (Wide) tasks are handled by the Driver for global synthesis and reasoning. This cycle repeats iteratively until the query is resolved.}
    \label{fig:model_framework}
\end{figure*}

\subsection{Query Decomposition via Directed Acyclic Graphs}
\label{subsec:dag}

While the distributed architecture described in \S\ref{subsec:method_arch} enables massive parallelization, it introduces a capability gap: lightweight Workers ($\theta_{\mathcal{W}}$) often lack the reasoning depth required to handle complex, multi-hop queries directly. To bridge this gap, we draw inspiration from distributed computing systems (e.g., Apache Spark) and utilize the Driver $\theta_{\mathcal{D}}$ as a central planner. The Driver decomposes the user query $q$ into a Directed Acyclic Graph (DAG) $\mathcal{G} = (\mathcal{V}, \mathcal{E})$, where nodes $v \in \mathcal{V}$ represent executable actions and edges denote data dependencies. To accommodate the disparate capabilities of the models, we classify nodes into two distinct functional types:

\paragraph{1. Narrow Actions (Map: Atomic Extraction).}
Narrow actions rely exclusively on local context and are executed by Workers. To mitigate the risk of reasoning failure in smaller models, we restrict these actions to \textbf{atomic, entity-centric extraction}. For a narrow action $v^{\text{N}}$ containing the instruction $i_v$, each Worker processes its persistent local shard $\vx^{(k)}$ independently:
\begin{equation}
    \vz_v^{(k)} \sim P_{\theta_{\mathcal{W}}}\left( \cdot \mid \vx^{(k)}, i_{v^{\text{N}}} \right).
\end{equation}
Here, the instruction $i_{v^{\text{N}}}$ is formulated not as a complex reasoning question, but as a direct lookup directive focusing on key entities or terms (e.g., ``Locate the birth date of entity A''). This constraint ensures that Workers operate purely as high-recall information scanners, effectively decoupling the extraction of evidence from the burden of logical deduction.

\paragraph{2. Wide Actions (Reduce: Logical Synthesis).}
Wide actions aggregate the outputs of preceding nodes to resolve cross-shard dependencies. These are executed by the Driver to perform \textbf{reasoning and synthesis}. For a wide action $v^{\text{W}}$ that depends on a set of parent outputs $\mathcal{Z}_{\text{parent}}$, the Driver generates a new insight:
\begin{equation}
    \vz_v^{\text{W}} = P_{\theta_{\mathcal{D}}}\left(\cdot \mid \mathcal{Z}_{\text{parent}}, v^{\text{W}}\right).
\end{equation}
Wide actions serve as the ``logic gates'' of the system. By processing only the distilled evidence $\mathcal{Z}_{\text{parent}}$ rather than the massive raw context, the Driver is able to perform multi-hop deduction and conflict resolution within a concise, high-fidelity window.

\subsection{Dynamic Orchestration: Iterative DAG Refinement}
\label{subsec:execution}

The optimal computation topology for a query is rarely known \textit{a priori}. \ourMethod{} therefore employs an \textbf{online planning process} formalized as a multi-stage loop: the Driver iteratively constructs, executes, and evaluates DAGs until the query is resolved.

\paragraph{1. Planning Phase (DAG Construction).}
At stage $t$, the Driver $\theta_{\mathcal{D}}$ acts as a Planner, conditioning on query $q$ and the execution history to generate a DAG $\mathcal{G}_t$. The graph mixes Narrow nodes (shard-level probes) and intermediate Wide nodes (partial synthesis), structured so that the \textbf{terminal node}---a Wide Action representing $q$ itself---receives the most refined information flow.

\paragraph{2. Execution Phase (Evidence Pooling).}
The system executes $\mathcal{G}_t$ in topological order, maintaining a stage-specific evidence pool $\mathcal{Z}_t$. \textbf{Narrow Actions} are dispatched to Workers $\theta_{\mathcal{W}}$, which scan their shards in parallel and materialize atomic facts into $\mathcal{Z}_t$. \textbf{Wide Actions} are then processed by the Driver as a Reducer: it retrieves dependent outputs from $\mathcal{Z}_t$, performs logical synthesis, and writes results back into the pool.

\paragraph{3. Evaluation and Refinement.}
Upon DAG completion, the Driver examines the terminal node's output $\hat{\vy}_t$. If the evidence suffices to answer $q$ with high confidence, the process terminates. Otherwise, the Driver rejects the output and advances to stage $t+1$, where it leverages $\mathcal{Z}_t$ to generate a refined graph $\mathcal{G}_{t+1}$ targeting the identified information gaps.

\subsection{Optimizing the Driver via Reinforcement Learning}
\label{subsec:rl}

Since optimal intermediate DAG structures lack ground truth supervision, we treat the Driver as an agent and optimize it using Reinforcement Learning. We employ \textbf{Group Relative Policy Optimization (GRPO)}~\citep{shao2024deepseekmath} to refine the Driver's policy. Unlike standard single-turn generation, a complete inference episode in \ourMethod{} produces a trajectory $\tau$ consisting of multiple discrete sequences $\tau = (s_1, \dots, s_T)$, corresponding to the iterative planning steps and the final reasoning output.

We optimize the policy against a composite reward function:
\begin{equation}
\label{eq:reward}
    R = \lambda_{\text{a}} R_{\text{acc}} + \lambda_{\text{e}} R_{\text{evd}} + \lambda_{\text{f}} R_{\text{fmt}}.
\end{equation}
The \textbf{Accuracy Reward} $R_{\text{acc}} \in \{0, 1\}$ evaluates the correctness of the final response $\vy$ against the ground truth. To ensure high-fidelity grounding, we introduce an \textbf{Evidence Reward} $R_{\text{evd}} \in \{0, 1\}$, which employs a judge model to verify that $\vy$ is logically derived strictly from the accumulated evidence $\mathcal{Z}$. This explicitly penalizes answers that are factually correct relative to external knowledge but unsupported by the retrieved context (hallucinations). Finally, to enforce structural validity, the \textbf{Format Reward} $R_{\text{fmt}} \in \{-1, 0\}$ applies a penalty if the generated DAG contains syntax errors or invalid dependencies, and is 0 otherwise.

A critical challenge in this multi-turn setting is credit assignment, as task-level utility is only observable at the terminal state. Consequently, following~\citet{yu2025memagent}, we \textbf{broadcast} the terminal rewards ($R_{\text{acc}}$ and $R_{\text{evd}}$) to all intermediate sequences $s_t$ within the trajectory. By propagating the final outcome back to the planning steps, we incentivize the Driver to construct efficient, precise retrieval plans that lead to correct reasoning, while $R_{\text{fmt}}$ provides immediate feedback at each step to ensure executability.

\begin{table*}[t]
\centering
\sisetup{detect-weight,mode=text}
\renewcommand{\bfseries}{\fontseries{b}\selectfont}
\caption{Evaluation results of models on long-context benchmarks. \textbf{LongBench v2} and \textbf{RULER-QA} numbers are accuracy. For \textbf{${\infty}$Bench}, we report MC (Multiple Choice), QA (Question Answering), and their average accuracy. \ourMethod{} (RL) represents our proposed method after GRPO training. See~\Cref{tab:cross_model} for more model families results.}
\label{tab:combined_results}
\renewcommand\arraystretch{1.05}
\resizebox{\linewidth}{!}{%
\begin{tabular}{
  l
  S[table-format=2.1] S[table-format=2.1] >{\columncolor{avgcol}}S[table-format=2.1]
  S[table-format=2.1] S[table-format=2.1] S[table-format=2.2] >{\columncolor{avgcol}}S[table-format=2.2]
  S[table-format=2.2] S[table-format=2.2] >{\columncolor{avgcol}}S[table-format=2.2]
}
\toprule
& \multicolumn{3}{c}{\textbf{LongBench v2}} & \multicolumn{4}{c}{\textbf{RULER-QA}} & \multicolumn{3}{c}{\textbf{$\infty$Bench}} \\
\cmidrule(lr){2-4} \cmidrule(lr){5-8} \cmidrule(lr){9-11}
\textbf{Method} & {\textbf{Medium}} & {\textbf{Long}} & \multicolumn{1}{>{\columncolor{avgcol}}c}{\textbf{Avg}} & {\textbf{256K}} & {\textbf{512K}} & {\textbf{1M}} & \multicolumn{1}{>{\columncolor{avgcol}}c}{\textbf{Avg}} & {\textbf{En.MC}} & {\textbf{En.QA}} & \multicolumn{1}{>{\columncolor{avgcol}}c}{\textbf{Avg}} \\
\midrule

\modelrow{Qwen3-8B (Thinking Mode)} \\
Full Long Context & 28.8 & 32.4 & 30.6 & {---} & {---} & {---} & {---} & 65.94 & 49.86 & 57.90 \\
RAG               & 30.8 & 32.1 & 31.5 & 52.4 & 33.5 & 10.9 & 32.27 & 66.35 & 53.15 & 59.75 \\
CoA               & 29.4 & 24.0 & 26.7 & 44.8 & 46.0 & 44.6 & 45.13 & 41.96 & 22.38 & 32.17 \\
LLM$\times$MapReduce     & 31.9 & 29.2 & 30.6 & 75.3 & 73.2 & 72.4 & 73.63 & 48.03 & 42.16 & 45.10 \\
RLM               & 9.3  & 10.2 & 9.8  & 11.4 & 8.5  & 0.5  & 6.80  & 13.53 & 17.09 & 15.31 \\
\midrule
\textbf{\ourMethod{}} & 34.4 & 36.1 & 35.3 & \bf 77.6 & \bf 77.5 & \bf 78.4 & \bf 77.83 & 69.25 & 52.14 & 60.70 \\
\textbf{\ourMethod{} (RL)} & \bf 36.3 & \bf 39.8 & \bf 38.1 & 76.5 & 77.0 & 76.5 & 76.67 & \bf 72.93 & \bf 54.42 & \bf 63.68 \\

\midrule

\modelrow{Qwen3-14B (Thinking Mode)} \\
Full Long Context & 42.3 & 38.9 & 40.6 & {---} & {---} & {---} & {---} & 68.56 & 58.12 & 63.34 \\
RAG               & 38.5 & 40.9 & 39.7 & 60.1 & 41.5 & 23.5 & 41.70 & 70.23 & 54.34 & 62.29 \\
CoA               & 23.3 & 36.1 & 29.7 & 53.5 & 44.6 & 42.3 & 46.80 & 53.71 & 34.38 & 44.05 \\
LLM$\times$MapReduce     & 31.1 & 33.7 & 32.4 & 75.6 & 75.9 & 75.5 & 75.67 & 51.96 & 45.58 & 48.77 \\
RLM               & 12.1 & 15.7 & 13.9 & 18.4 & 11.5 & 1.4  & 10.43 & 17.40 & 20.80 & 19.10 \\
\midrule
\textbf{\ourMethod{}} & 41.8 & 48.7 & 45.3 & 78.7 & 76.5 & 77.9 & 77.71 & 72.93 & 56.49 & 64.71 \\
\textbf{\ourMethod{} (RL)} & \bf 43.7 & \bf 50.9 & \bf 47.3 & \bf 79.2 & \bf 78.8 & \bf 77.0 & \bf 78.33 & \bf 75.11 & \bf 59.26 & \bf 67.19 \\

\bottomrule
\end{tabular}
}
\end{table*}

\section{Experimental Setup}

\subsection{Evaluation Protocol}

\paragraph{Baselines.}
We compare \ourMethod{} against representative methods covering direct processing, retrieval, and agentic workflows: (1) \textbf{Naive Long-Context}, where the full context is fed directly into the model's window to assess native performance and degradation; (2) \textbf{Standard RAG}, implementing a dense retrieval pipeline using Qwen3-Embedding-4B to fetch the top-16 chunks (256 tokens each) based on cosine similarity; (3) \textbf{Chain-of-Agents (CoA)} \citep{zhang2024chain}, which processes partitioned context sequentially, passing summarized information between worker agents; (4) \textbf{LLM$\times$MapReduce} \citep{zhou2024llm}, a divide-and-conquer approach employing parallel evidence extraction followed by a single aggregation step;
(5) \textbf{Recursive Language Model (RLM)} \citep{zhang2025recursive}, an interactive system where the model recursively queries the text environment to gather information. (See more details in~\Cref{subsec:eval_details})

\paragraph{Benchmarks.}
We assess performance across three benchmarks: (1) \textbf{LongBench v2}~\citep{bai2024longbench}, a comprehensive multi-choice suite evaluating contexts ranging from 8K to 2M tokens. We evaluate on the Medium (32K-128K) and Long ($>$128K) subsets. (2) \textbf{RULER-QA}~\citep{hsieh2024ruler}, a synthetic benchmark testing multi-hop reasoning over arbitrary context length. We focus on this QA task with the lengths of 256K, 512K, and 1M tokens. (3) \textbf{$\infty$Bench}~\citep{zhang2024inftybench}. We evaluate all models on two tasks in this benchmark:  long-book question answering (En.QA), and multi-choice question-answering (En.MC). The average evaluation length is from 150K to 200K.

\paragraph{Models and Implementation.}
We evaluate \ourMethod{} and the state-of-the-art baselines using three backbone LLMs: \textbf{Qwen3-8B}, \textbf{Qwen3-14B}, and \textbf{Gemini-3-Pro-Preview}. For open-weight models, we utilize SGLang~\citep{Zheng2023SGLangEE} for efficient inference, with context length extended to 128K by YaRN~\citep{peng2023yarn} across all baselines. In~\ourMethod{}, we use the Qwen3-4B-Instruct-2507 as the Worker LLM across evaluations except specifically mentioned. We set the context shard size of each worker as 4096 and  maximum replanning stages as 10.

\subsection{Implementation Details.}

\paragraph{Training Data Construction.}
We construct our training set using the \textbf{MuSiQue} dataset~\citep{Trivedi2021MM} to ensure high reasoning complexity. To synthesize the massive contexts required to test distributed scaling (16K--512K tokens), we developed a \textbf{Recursive Context Augmentation} pipeline. Leveraging {Qwen3-235B-A22B-Think-2507}, we identify entity mentions $e$ within supporting claims and recursively replace them with the full content of their linked Wikipedia pages ($P_e$). This transformation follows the schema $\text{``}\dots [e] \dots\text{''} \to \text{``}\dots [\textsc{Content}(P_e)] \dots\text{''}$, effectively embedding sparse retrieval targets within extensive unstructured text. We apply a consistency filter to prune samples where expansion introduces logical contradictions, yielding 4.5k high-fidelity samples. More details are listed in~\Cref{subsec:data_details}.

\paragraph{Training Setup.}
We train two models, Qwen3-8B (thinking) and Qwen3-14B (thinking), employing Group Relative Policy Optimization (GRPO)~\citep{shao2024deepseekmath}. Training proceeds for 2 epochs on the curated 4.5K dataset using the AdamW optimizer with a constant learning rate of $1\mathrm{e}{-6}$ and a 5-step linear warmup. During the rollout phase, we set a global batch size of 256 and sample $G=8$ trajectories per prompt, enforcing a maximum context window of 24K tokens and a generation limit of 16K tokens. We use Qwen3-4B-Instruct-2507 as the Worker LLM (shard size=4096) during training. We utilize Qwen3-235B-A22B-Instruct-2507 as the judge model for reward computation. The weighting coefficients for the composite reward function (\Cref{eq:reward}) are set to $\lambda_{\text{acc}}=1.0$ and $\lambda_{\text{fmt}}=\lambda_{\text{evd}}=0.5$.

\section{Results and Analyses}

\subsection{Main Results}
\label{subsec:main_results}

We validate \ourMethod{} on LongBench v2, RULER-QA, and $\infty$Bench, using Qwen3-8B, Qwen3-14B, and Gemini-3-Pro. The results confirm the effectiveness of our three core architectural pillars: distributed grounding, decoupled reasoning, and economic scalability.

\paragraph{Distributed Partitioning Ensures Robust Grounding at Scale.}
In Table~\ref{tab:combined_results}, our results demonstrate that by utilizing distributed context partitioning, \ourMethod{} achieves superior contextual grounding compared to all baselines, maintaining robustness even up to the 1M token scale. This is strikingly evident on \textbf{RULER-QA}: while RAG fails to retrieve relevant segments from the massive search space—collapsing to 10.9\% accuracy with Qwen3-8B—\ourMethod{} sustains \textbf{78.4\%}. Unlike methods that degrade with length, our distributed architecture renders grounding performance invariant to the total context size. By unshackling the system from the bottlenecks of heuristic retrieval and static windowing, \ourMethod{} outperforms the strongest baseline (LLM$\times$MapReduce) by 6\%, proving that dynamic partitioning is essential for high recall in the million-token regime.

\begin{figure}[t]
    \centering
    \includegraphics[width=\linewidth]{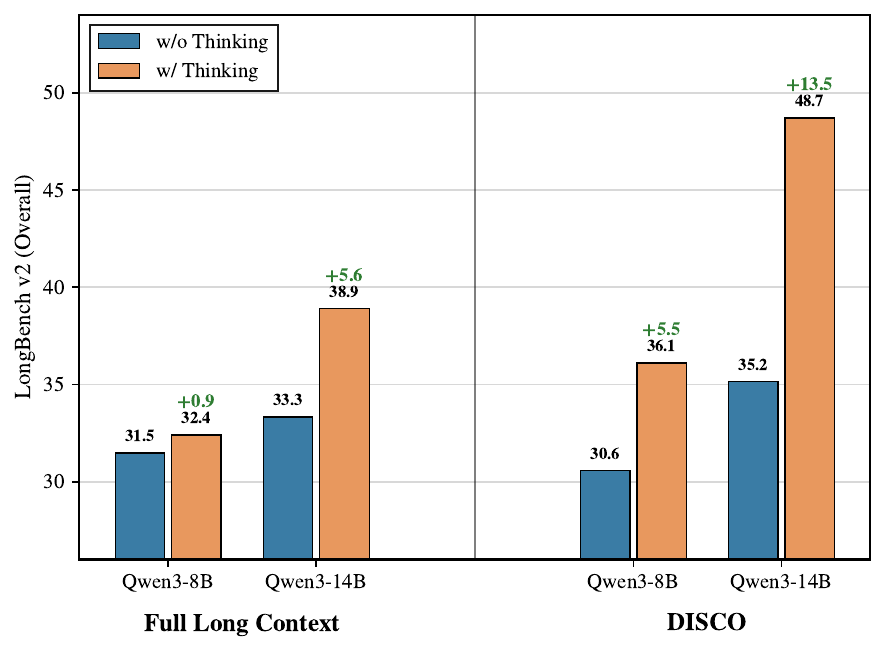}
    \caption{Thinking Mode accuracy gains on LongBench v2. \ourMethod{} amplifies the benefit of extended reasoning (e.g., +13.5 vs.\ +5.6 for Qwen3-14B) by freeing the Driver from grounding overhead.}
    \label{fig:thinking}
\end{figure}

\paragraph{Decoupling and Replanning Unleash Reasoning Potential.}
We attribute the superior performance of \ourMethod{} to the Driver's dynamic DAG planning, which structurally decouples \textit{grounding} (Workers) from \textit{reasoning} (Driver). Unlike \textbf{LLM$\times$MapReduce}—which relies on a rigid, single-pass scan where workers must handle full query complexity—\ourMethod{} operates iteratively: the Driver decomposes multi-hop queries into atomic retrieval steps (Narrow Actions) and synthesizes intermediate results (Wide Actions), dynamically re-planning based on feedback. On \textbf{LongBench v2}, \ourMethod{} (Qwen3-14B) achieves a \textbf{9.8 point} gain over Full Long Context (48.7\% vs 38.9\%) and substantially outperforms LLM$\times$MapReduce (33.7\%).

This decoupling also amplifies intrinsic reasoning capacity. As shown in~\Cref{fig:thinking}, enabling Thinking Mode under Full Long Context yields modest gains (+0.9 for 8B, +5.6 for 14B), as extended reasoning is largely consumed by grounding through noisy context. \ourMethod{} magnifies these gains to +5.5 and +13.5 respectively—since Workers have already distilled compact evidence, the Driver's thinking budget is spent entirely on reasoning.

\begin{figure}[t]
    \centering
    \includegraphics[width=\linewidth]{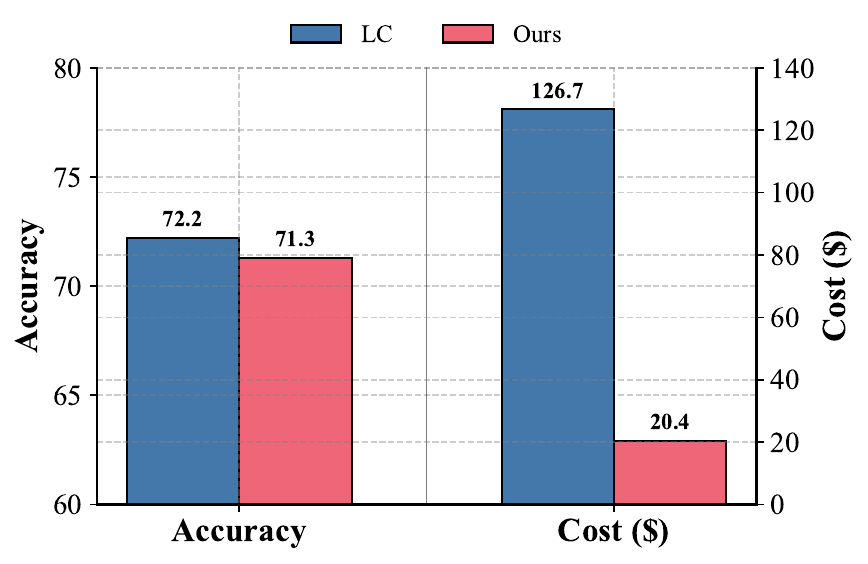}
    \caption{The accuracy and evaluation costs on LongBench v2 (Long) using Gemini-3-Pro as the Driver. \ourMethod{} matches the performance of full context processing while reducing financial cost by over 80\%.}
    \label{fig:gemini}
\end{figure}

\paragraph{Architecture Decouples Cost from Scale.}
Finally, \ourMethod{} proves effective at enabling the economic deployment of frontier-class reasoning on massive context. By offloading token-intensive scanning to lightweight Workers, we decouple the cost of \textit{reading} from the cost of \textit{thinking}. As shown in ~\Cref{fig:gemini} with Gemini-3-Pro, \ourMethod{} retains the reasoning parity of the Full Context baseline (71.3\% vs. 72.2\%) while slashing evaluation costs by \textbf{6.2$\times$} (\$126.7 $\to$ \$20.4). This validates \ourMethod{} as a scalable wrapper that allows superior LLMs to process million-token contexts without the prohibitive quadratic costs of standard attention.

\begin{figure}[t]
    \centering
    \includegraphics[width=\linewidth]{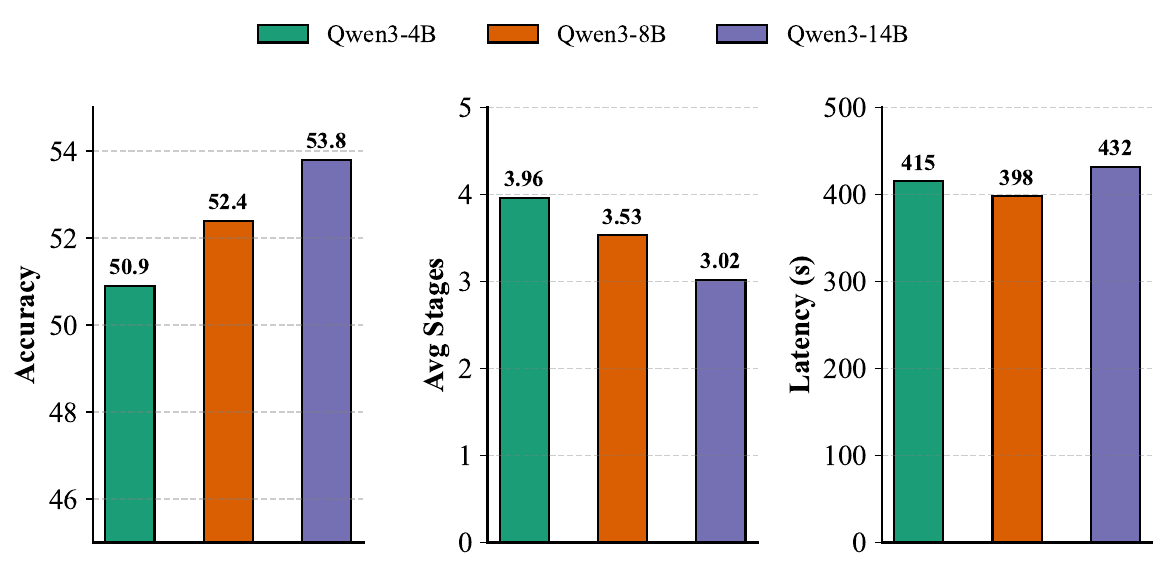}
    \caption{Impact of Worker model scale on accuracy and planning efficiency. While larger Workers yield only marginal accuracy gains, they significantly reduce the number of planning stages required by the Driver to converge.}
    \label{fig:worker_scale}
\end{figure}

\subsection{Impact of Worker LLMs}

\paragraph{Sensitivity to Worker Model Scale.}
We evaluate \ourMethod{} with different Worker sizes (Qwen3-4B, Qwen3-8B, and Qwen3-14B in non-thinking mode) while fixing the Driver (Qwen3-14B after RL training). As shown in ~\Cref{fig:worker_scale}, increasing Worker scale yields 2.9\% accuracy gains (50.9\% $\to$ 53.8\%), which are significantly lower than scaling the Driver (39.8\% $\to$ 50.9\% for 8B to 14B.). This confirms that \emph{contextual grounding} is less parameter-sensitive than \emph{global reasoning}. In addition, we observe that larger Workers significantly improve planning efficiency: the average execution stages drop from \textbf{3.96} (4B) to \textbf{3.02} (14B). Crucially, this reduction in search depth offsets the increased per-token cost of the larger model, resulting in comparable total latency (415s vs. 432s).

\begin{table}[t]
    \centering
    \caption{\textbf{Ablation of Reward Design} on LongBench v2 (Qwen3-8B). \emph{Latency} includes retries from format errors.}
    \label{tab:reward_ablation}
    \resizebox{\linewidth}{!}{
    \begin{tabular}{l c c c}
        \toprule
        \textbf{Configuration} & \textbf{Acc (\%)} & \textbf{Avg Stages} & \textbf{Latency (s)} \\
        \midrule
        \textbf{Full Reward} & \textbf{39.8} & \textbf{3.39} & \textbf{381} \\
        \quad w/o $R_{\text{evd}}$ & 36.3 & 4.45 & 432 \\
        \quad w/o $R_{\text{fmt}}$ & 34.4 & 3.84 & 499 \\
        \bottomrule
    \end{tabular}
    }
\end{table}

\subsection{Impact of Reward Design}
\label{subsec:reward_ablation}

To validate our composite reward function (Eq.~\ref{eq:reward}), we ablate the Evidence ($R_{\text{evd}}$) and Format ($R_{\text{fmt}}$) rewards on Qwen3-8B (Table~\ref{tab:reward_ablation}).

\paragraph{Planning Efficiency ($R_{\text{evd}}$).}
Removing $R_{\text{evd}}$ reduces accuracy and spikes execution length. Without penalties for ungrounded hallucinations, the Driver fails to distinguish valid evidence from noise, leading to partially supported answers, frequent self-evaluation rejections, and costly re-planning loops.

\paragraph{Structural Stability ($R_{\text{fmt}}$).}
Omitting $R_{\text{fmt}}$ yields the lowest accuracy and highest latency. Without structural penalties enforcing schema compliance, the Driver frequently generates syntactically invalid DAGs. This triggers automatic retries, wasting computational cycles on exception handling rather than task progression.

\begin{figure}[t]
    \centering
    \includegraphics[width=\linewidth]{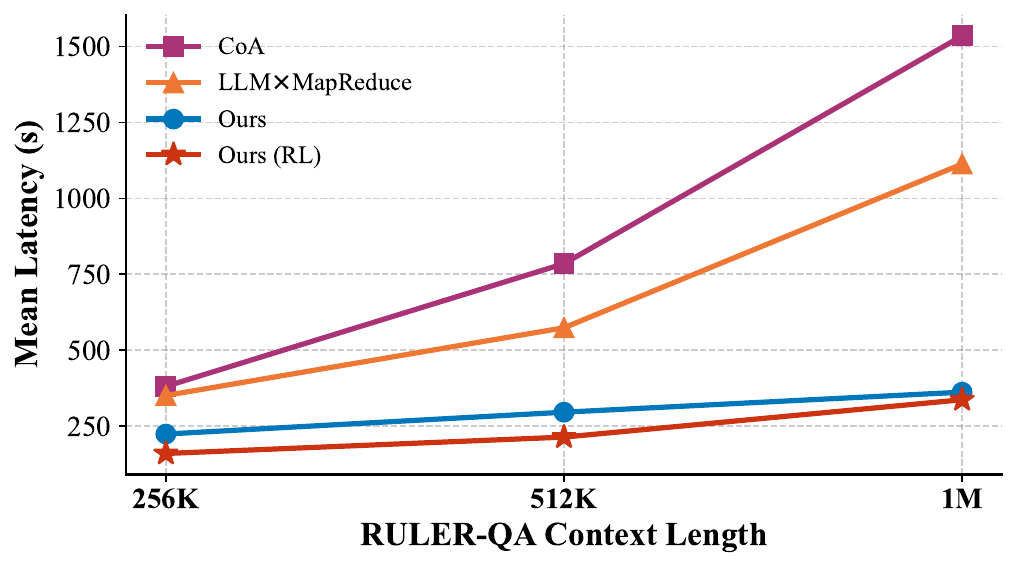}
    \caption{End-to-end latency on RULER-QA under an iso-compute budget (8$\times$H100). \ourMethod{} maintains near-constant latency as context scales to 1M tokens.}
    \label{fig:latency}
\end{figure}

\subsection{Inference Efficiency and Scalability}
We evaluate inference latency on the RULER-QA benchmark across context lengths of 256K to 1M under a fixed iso-compute budget of 8$\times$H100 GPUs. Baselines deploy a monolithic Qwen3-14B (TP=8), whereas \ourMethod{} adopts a heterogeneous strategy, allocating 4 GPUs to Driver (Qwen3-14B) and 4 to parallel Workers (Qwen3-4B). As shown in Figure~\ref{fig:latency}, \ourMethod{} significantly outperforms agentic baselines (CoA, LLM$\times$MapReduce), maintaining a near-flat latency profile while baselines exhibit steep linear growth. By offloading the high-volume token processing to efficient Workers, we circumvent the bottleneck of processing massive sequences with large parameters. Furthermore, the RL-tuned variant achieves the highest efficiency by optimizing planning paths to minimize execution steps. These results demonstrate the superior \textbf{long-context scaling potential} of \ourMethod{}.

\section{Related Work}

\paragraph{Distributed Computing Paradigms.}
The management of massive datasets was revolutionized by MapReduce~\citep{dean2004mapreduce} and Hadoop~\citep{hadoop}, which introduced a paradigm of moving computation to data rather than transferring data to compute nodes. Apache Spark further advanced this by introducing Resilient Distributed Datasets (RDDs) and Directed Acyclic Graph (DAG) scheduling, enabling efficient in-memory iterative processing. These systems decompose complex queries into \textit{Map} (parallel processing) and \textit{Reduce} (aggregation) operations. Our work adapts these classical distributed primitives to the domain of LLMs, treating long context as distributed data that requires orchestrated, parallel processing rather than monolithic ingestion.

\paragraph{Long-Context Processing in LLMs.}
Current approaches to handling extensive context generally fall into three categories: \textbf{(1) Architectural Extrapolation.} 
A primary line of research focuses on extending the context window of single models via length extrapolation strategies~\citep{chen2023extending, peng2023yarn, xiong2023effective, an2024trainingfree} or memory-augmented architectures~\citep{behrouz2024titans, behrouz2025s} empowered by test-time training~\citep{tandon2025end}. While these methods expand the nominal window size, they frequently suffer from ``context rot,'' where reasoning capabilities degrade significantly as input length increases~\citep{hong2025context, modarressi2025nolima}. Furthermore, training memory modules from scratch poses generalization challenges that remain underexplored.
\textbf{(2) Retrieval and Agentic Memory.} 
To bypass architectural limits, systems rely on RAG or, more recently, agentic memory frameworks that selectively compress, edit, and retrieve context~\citep{chhikara2025mem0, li2025memos, zhou2025mem1, yan2025general}. While scalable, these select-based pipelines face a fundamental retrieval bottleneck: they optimize for local semantic similarity, often discarding implicitly pertinent information required for global reasoning~\citep{hu2025memory, du2025rethinking}. Even just-in-time mechanisms~\citep{yan2025general} ultimately rely on imperfect recall to recover lost information.
\textbf{(3) Agentic Workflows.} 
Recent works have proposed decomposing long context processing into agent workflows. Chain-of-Agents~\citep{zhang2024chain} and subsequent variants~\citep{qian2024long, zhao2024longagent} process chunks sequentially, passing information in a chain. However, this serialization incurs prohibitive latency and suffers from cascading information loss. LLM$\times$MapReduce~\citep{zhou2024llm} attempts parallel scanning evidences and aggregation using a single LLM for each user query, which often lacks the scalability to handle complex, multi-hop reasoning over millions of tokens.  While Recursive LMs~\citep{zhang2025recursive} introduce environment-based interaction for long context, they only remain effective upon massive LLMs (hundreds of billions of parameters) with strong coding capabilities.

In contrast, \ourMethod{} decouples contextual grounding from reasoning. By adopting a distributed system view, we enable smaller Workers to process short chunks in parallel (Map) while a central Driver orchestrates high-level logic (Shuffle/Reduce), overcoming the latency of sequential agents and the precision loss of retrieval systems, hence becoming more scalable for long context.

\section{Conclusion}
\label{sec:conclusion}

In this work, we reframe long-context scaling as a distributed computing problem. We introduce \ourMethod{}, a framework that structurally decouples \textit{contextual grounding} from \textit{logical reasoning}. By partitioning massive contexts across a fleet of Worker LLMs orchestrated by a central, GRPO-trained Driver, \ourMethod{} effectively eliminates context rot, maintaining robust fidelity up to one million tokens where monolithic architectures collapse.

Isolating the reasoning engine from raw data noise creates a high-signal environment optimized for complex, multi-hop tasks. This disaggregation not only elevates reasoning capabilities but also achieves the accuracy of frontier models at a fraction of the inference cost. Ultimately, \ourMethod{} signals a critical paradigm shift: moving away from passively expanding single-node context windows, and toward building active, distributed systems designed to navigate unbounded information landscapes.

\bibliography{reference}

\appendix
\section{Appendix}

\subsection{Algorithm}

\begin{algorithm}[!ht]
\small
\caption{Inference Dynamics of \ourMethod{}}
\label{alg:distributed_inference}
\begin{algorithmic}[1]
\REQUIRE Long context $\vx$, Query $q$, Driver Model $\theta_{\mathcal{D}}$, Worker Model $\theta_{\mathcal{W}}$.
\ENSURE Final Answer $\hat{\vy}$.
\STATE \textbf{Initialization:}
\STATE \quad Partition $\vx \to \{ \vx^{(1)}, \dots, \vx^{(K)} \}$ and persist to Workers.
\STATE \quad Initialize evidence pool: $\mathcal{Z}_0 \leftarrow \emptyset$.
\STATE \quad Initialize DAG state: $\mathcal{G}_0 \leftarrow (\mathcal{V}=\{q\}, \mathcal{E}=\emptyset)$.

\FOR{stage $t = 1$ to $T_{\max}$}
    \STATE \textcolor{gray}{\emph{// Phase 1: Planning \& Evaluation (Driver)}}
    \STATE Condition on query $q$ and history $\mathcal{Z}_{t-1}$.
    \STATE Generate frontier actions: $\mathcal{A}_t \sim \pi_{\text{plan}}(\cdot \mid q, \mathcal{G}_{t-1}, \mathcal{Z}_{t-1}; \theta_{\mathcal{D}})$.
    
    \IF{Action $\textsc{Answer}(\hat{\vy}) \in \mathcal{A}_t$}
        \STATE \textcolor{gray}{\emph{// Terminal state reached with sufficient confidence}}
        \STATE \textbf{return} $\hat{\vy}$.
    \ENDIF

    \STATE \textcolor{gray}{\emph{// Phase 2: Map (Narrow Actions)}}
    \STATE Identify Narrow actions: $\mathcal{A}_{\text{map}} = \{ (k, i_v) \in \mathcal{A}_t \}$.
    \FOR{each action $(k, i_v) \in \mathcal{A}_{\text{map}}$ \textbf{in parallel}}
        \STATE \textcolor{gray}{\emph{// Worker scans local shard for atomic facts}}
        \STATE Extract local evidence: $\vz^{(k)}_v \sim P_{\theta_{\mathcal{W}}}(\cdot \mid \vx^{(k)}, i_v)$.
        \STATE Add to evidence pool: $\mathcal{Z}_t \leftarrow \mathcal{Z}_{t-1} \cup \{ \vz^{(k)}_v \}$.
    \ENDFOR

    \STATE \textcolor{gray}{\emph{// Phase 3: Reduce (Wide Actions)}}
    \STATE Identify Wide actions: $\mathcal{A}_{\text{reduce}} = \{ (\mathcal{P}, i_v) \in \mathcal{A}_t \}$.
    \FOR{each action $(\mathcal{P}, i_v) \in \mathcal{A}_{\text{reduce}}$}
        \STATE \textcolor{gray}{\emph{// Driver synthesizes insights from parent outputs}}
        \STATE Gather parent artifacts: $\mathcal{Z}_{\text{parent}} \leftarrow \{ \vz_p \mid p \in \mathcal{P} \} \subseteq \mathcal{Z}_t$.
        \STATE Synthesize global insight: $\vz^{W}_v \sim P_{\theta_{\mathcal{D}}}(\cdot \mid \mathcal{Z}_{\text{parent}}, i_v)$.
        \STATE Update evidence pool: $\mathcal{Z}_t \leftarrow \mathcal{Z}_t \cup \{ \vz^{W}_v \}$.
    \ENDFOR

    \STATE Update Graph State: $\mathcal{G}_t \leftarrow \mathcal{G}_{t-1} \cup \mathcal{A}_t$.
\ENDFOR
\STATE \textbf{return} Best-effort response from final state $\mathcal{Z}_{T_{\max}}$.
\end{algorithmic}
\end{algorithm}

\subsection{Evaluation Details}
\label{subsec:eval_details}
\paragraph{Baselines.} We evaluate all baselines following their official implementations, with the exception of Chain-of-Agents, for which code is unavailable; in this case, we reproduced the method following the paper's details. We set the chunk size to 4096 for both Chain-of-Agents and LLM$\times$MapReduce. For the Recursive Language Model (RLM), we adopt the official settings with a maximum depth of 1 and 30 iterations. We observed that RLM fails to perform effectively with Qwen3-8B/14B. This aligns with the original authors' findings that smaller LLMs lack the sufficient coding and planning capabilities required for recursive interaction within REPL environments. For the full long context baseline, we truncate the length within 128K in the middle following LongBench v2 and $\infty$Bench official settings.

\paragraph{Benchmarks.} We found that the standard F1 score metric for En.QA is highly unstable due to the diversity of prediction formats. To address this, we employ Qwen3-235B-A22B-Instruct as an external judge to verify if the prediction semantically matches the ground truth, assigning a binary score of 0 or 1.

\subsection{Training Data Construction}
\label{subsec:data_details}
We derive our training set from the \textbf{MuSiQue} dataset~\citep{Trivedi2021MM}, selected for its high complexity in multi-hop reasoning. However, the native context length of MuSiQue (comprising limited Wikipedia paragraphs) is insufficient to saturate the capacity of our distributed Worker fleet or test long-context robustness. To address this, we developed a \textbf{Recursive Context Augmentation} pipeline to scale the input length from standard baselines to 16K--512K tokens.

We employ {Qwen3-235B-A22B-Think-2507} to traverse the entity links within the original supporting facts. For a given claim $C$ containing an entity mention $e$ that links to an external Wikipedia page $P_e$, we perform a semantic expansion. We define a transformation function $\mathcal{T}(C, e, P_e)$ that rewrites the claim to embed the full content of the external page while maintaining narrative coherence. The expansion follows the template:
\begin{equation}
\begin{aligned}
    \mathcal{T}: \quad & \text{``} \dots \text{ [} e \text{] } \dots \text{''} \\
    & \longrightarrow \quad \text{``} \dots \text{ [} \textsc{Content}(P_e) \text{] } \dots \text{''}
\end{aligned}
\end{equation}
\noindent \textit{Representative Case:} Consider a claim mentioning a location.
\begin{itemize}
    \item \textit{Original Claim:} ``The event took place in \textbf{[Paris]}, spanning three days.''
    \item \textit{Expanded Context:} ``The event took place in \textbf{[Paris, which is the capital and most populous city of France... (full text of page) ...]}, spanning three days.''
\end{itemize}
This method effectively creates massive context blocks that are semantically linked to the query but dispersed, forcing the model to perform long-range information retrieval rather than local scanning.

To ensure data integrity, we apply a \textbf{Consistency Filter}. For every expanded sample, we feed the augmented context and the original ground truth reasoning path to {Qwen3-235B-A22B-Think-2507}. The model filters out any samples where the injected external text introduces factual contradictions or invalidates the original reasoning chain. The final dataset consists of \textbf{4.5k samples} with a length distribution uniformly spread between 16K and 512K tokens.

\subsection{More Experimental Results}

\begin{figure}[t]
    \centering
    \includegraphics[width=\linewidth]{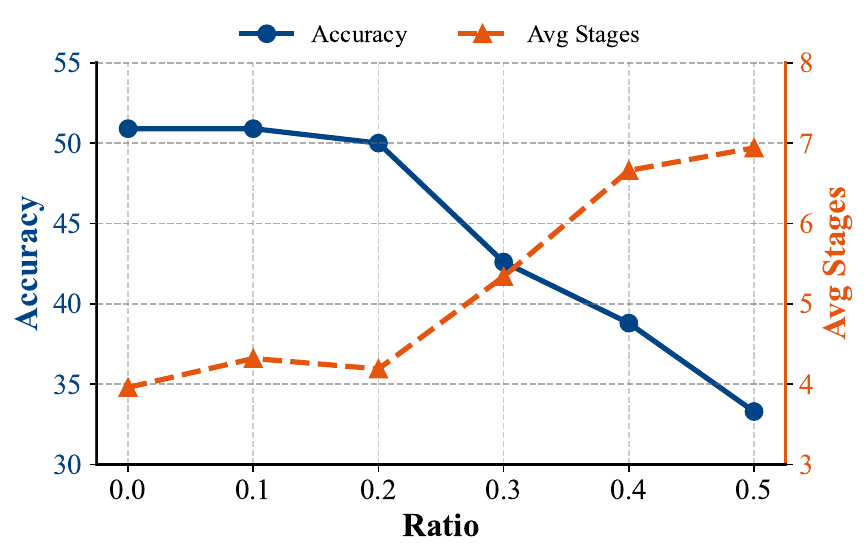}
    \caption{System resilience to information loss (Fact Dropout). Performance remains stable under moderate noise ($p \le 0.2$) but degrades sharply as excessive loss forces the Driver into exhaustion loops ($p \ge 0.3$).}
    \label{fig:robustness}
\end{figure}

\begin{table}[t]
    \centering
    \caption{Comparison of \ourMethod{} using a Generative Worker versus a Dense Retriever baseline. The Driver is Qwen3-14B after our RL training.}
    \label{tab:ablation_worker}
    \renewcommand{\arraystretch}{1.2}
    \setlength{\tabcolsep}{4pt}
    \begin{tabular}{@{}l ccc@{}}
        \toprule
        & \multicolumn{3}{c}{\textbf{LongBench v2}} \\
        \cmidrule(lr){2-4}
        \textbf{Worker Type} & \textbf{Medium} & \textbf{Long} & \textbf{Avg} \\
        \midrule
        Qwen3-Embedding-4B & 37.2 & 41.9 & 39.6 \\
        Qwen3-4B-Instruct  & \textbf{43.7} & \textbf{50.9} & \textbf{47.30} \\
        \bottomrule
    \end{tabular}
\end{table}

\paragraph{Fault Tolerance via Iterative Replanning.}
To evaluate resilience against extraction failures, we simulate a ``Fact Dropout'' scenario where fraction $p \in [0.0, 0.5]$ of narrow action outputs are discarded. As shown in Figure~\ref{fig:robustness}, \ourMethod{} exhibits a distinct phase transition. In the low-noise regime ($p \le 0.2$), the system is highly resilient: accuracy remains stable near 51\% with negligible increase in planning steps ($\approx$4 stages). This indicates the Driver's reasoning is robust to bridge partial information gaps without incurring overhead. However, a tipping point occurs at $p=0.3$: average stages spike sharply (from $\sim$4 to $\sim$7) as Driver is forced into repeated search loops, causing accuracy to plummet to 33\% at $p=0.5$. This confirms that while \ourMethod{} withstands moderate worker failure, excessive information loss eventually exhausts the inference budget ($T_{\max}=10$), leading to termination before convergence.

\paragraph{Generative Worker LLM Outperforms Dense Retriever.}
A core premise of \ourMethod{} is that contextual grounding requires \emph{processing} information using generative LLMs, not just matching it. To validate the necessity of generative Worker LLMs, we conducted an ablation study comparing our distributed active processing approach against a standard passive retrieval baseline. We replaced the generative Worker (Qwen3-4B-Instruct) with a dense retriever of equivalent parameter scale (Qwen3-Embedding-4B). In the baseline setting, instead of executing Map operations, the system retrieves the top-$k$ (where $k=16$, chunk size $=256$) most semantically similar chunks based on the queries from Planner and feeds them directly to the Reducer. Empirical results in~\Cref{tab:ablation_worker} reveal a substantial performance margin: \ourMethod{} achieves a score of \textbf{47.3}, significantly outperforming the embedding-based baseline at \textbf{39.6} on LongBench v2. This confirms that standard RAG pipelines suffer from a ``semantic gap''---embedding similarity often fails to capture diverse and task-specific dependencies required for effective grounding. In contrast, the generative Worker acts as a dynamic filter, performing inference-time computation to extract implicitly relevant evidence $\vz^{(k)}$ that a static vector index misses.

\begin{table}[t]
    \centering
    \caption{Generalization of \ourMethod{} across diverse Driver and Worker model combinations on \textbf{LongBench v2 (Long)}. \ourMethod{} consistently improves over the Full-Context CoT baseline regardless of the driver model family.}
    \label{tab:cross_model}
    \renewcommand{\arraystretch}{1.2}
    \setlength{\tabcolsep}{4pt}
    \resizebox{\linewidth}{!}{
    \begin{tabular}{@{}l l c@{}}
        \toprule
        \textbf{Driver} & \textbf{Worker} & \textbf{Score} \\
        \midrule
        \multirow{3}{*}{LLaMA-3.1-70B-Instruct}
            & --- (Full Context, CoT)        & 25.90 \\
            & Ministral-3-8B-Instruct (FP8)  & \textbf{28.70} \\
            & Qwen3-4B-Instruct              & 27.80 \\
        \cmidrule(lr){1-3}
        \multirow{2}{*}{Ministral-8B-Reasoning}
            & --- (Full Context, CoT)        & 32.41 \\
            & Qwen3-4B-Instruct-2507         & \textbf{37.96} \\
        \cmidrule(lr){1-3}
        Gemini-3-Pro-Preview
            & Ministral-3-8B-Instruct (FP8)  & \textbf{70.37} \\
        \bottomrule
    \end{tabular}
    }
\end{table}

\paragraph{Generalization across Driver and Worker Families.}
To verify that the benefits of \ourMethod{} are not specific to the Qwen3 model family used in \Cref{tab:combined_results}, we further evaluate diverse Driver--Worker pairings on LongBench v2 (Long). As shown in~\Cref{tab:cross_model}, the gain pattern is preserved across families: pairing LLaMA-3.1-70B-Instruct with a Ministral-3-8B Worker lifts accuracy from \textbf{25.9} (Full-Context CoT) to \textbf{28.7} (\textbf{+2.8}), and Ministral-8B-Reasoning improves from \textbf{32.41} to \textbf{37.96} (\textbf{+5.55}) when paired with a Qwen3-4B-Instruct-2507 Worker. The Worker family itself is fungible---LLaMA-3.1-70B with Ministral-3-8B and Qwen3-4B Workers reaches comparable scores (28.7 vs.\ 27.8)---indicating that the Driver only requires \emph{atomic factual extraction}, not a specific model lineage. Finally, Gemini-3-Pro-Preview with a Ministral-3-8B Worker reaches \textbf{70.37}, within $\sim$1 point of the Qwen3-4B Worker result reported in \Cref{fig:gemini} (71.3). Together, these results corroborate the decoupling thesis: with grounding distributed to lightweight, swappable Workers, the system's reasoning ceiling is dictated by the Driver, while the Worker only needs to be a competent extractor.

\subsection{Case Studies of RL Benefits}
\label{subsec:rl_case_studies}

To complement the quantitative results, we trace representative \ourMethod{} trajectories on LongBench v2 (Long) under two Driver configurations: \textbf{After RL} (Qwen3-8B-\ourMethod{}, GRPO-tuned) and \textbf{Before RL} (Qwen3-8B base), with Workers held constant so the only variable is the Driver's planning capability. Two qualitative improvements consistently emerge across the 15 sampled trajectories, illustrated by the case boxes below.

\paragraph{Atomic, Entity-Specific Query Decomposition.}
Before RL, the Driver routinely forwards generic, near-verbatim multi-hop queries to Workers, often re-using the same shard key across multiple NARROW transforms---diluting evidence quality and forcing extra replanning rounds. After RL, the Driver decomposes the question into atomic sub-queries that each target a distinct entity localized in a single shard. Across the suite, this learned decomposition reduces the average planning stages to \textbf{3.39} (\Cref{tab:reward_ablation}), confirming that the Driver learns to atomize multi-hop queries rather than offloading them wholesale.

\begin{casebox}{Case A: Kalamang Language Translation}
\textbf{Task.} Translate the Kalamang phrase ``\textit{terus ter-nan koyet inier tamu kon misis wis}'' into English.

\smallskip
\textbf{After RL.} \emph{317 facts, converged in 1 stage.} The Driver issues three NARROWs that partition the phrase into disjoint word-level segments:
\begin{quote}\footnotesize\ttfamily\raggedright
N1: key=``terus ter-nan'',\ \ \ \,query=``Extract English translations for `terus', `ter-nan'\,''\\
N2: key=``koyet inier tamu'', query=``Extract English translations for `koyet', `inier', `tamu'\,''\\
N3: key=``kon misis wis'',\ \ \ \,query=``Extract English translations for `kon', `misis', `wis'\,''
\end{quote}

\textbf{Before RL.} \emph{69 facts ($4.6\times$ fewer), 1 stage.} The Driver reuses one document-level key for all four NARROWs, and one transform pastes the entire user query verbatim into a single Worker prompt:
\begin{quote}\footnotesize\ttfamily\raggedright
N1--N3: key=``kalamang language'' (reused), query=``Translate `koyet' / `wis' / `misis'\,''\\
N4:\ \ \ \ \,key=``kalamang language'', query=``Translate the phrase `terus ter-nan koyet inier tamu kon misis wis'\,''
\end{quote}

\textit{Observation.} The After-RL Driver assigns disjoint, atomic key segments aligned to chunked storage. The Before-RL Driver collapses to one generic key and pastes the entire user query into N4---an instruction that no single Worker chunk can satisfy.
\end{casebox}

\begin{casebox}{Case D: Zhuang-to-Chinese Translation}
\textbf{Task.} Translate the Zhuang sentence ``\textit{Gou sien youq aen hekdiemq}'' into Chinese using a provided textbook.

\smallskip
\textbf{After RL.} \emph{37 facts, 1 stage.} The Driver uses the exact source sentence as the extraction key, directing Workers to the precise location of its translation:
\begin{quote}\footnotesize\ttfamily\raggedright
N1: key=``Gou sien youq aen hekdiemq'', query=``Find the exact Chinese translation for this Zhuang sentence''\\
N2: key=``Zhuang vocabulary'',\ \ \ \ \ \ \ \ \ \ \ \,query=``Extract Chinese meanings for related Zhuang words''
\end{quote}

\textbf{Before RL.} \emph{2 facts ($\mathbf{18.5\times}$ fewer), 5 stages, force-finalized.} All NARROWs reuse a generic topic label that does not match any chunk substring:
\begin{quote}\footnotesize\ttfamily\raggedright
N1--N3: key=``Zhuang language translation'' (reused, topic label, no chunk match)
\end{quote}

\textit{Observation.} The After-RL Driver localizes against the unique source sentence; the Before-RL Driver's topic label produces no surface-level matches and the run exhausts $T_{\max}{=}5$ without converging.
\end{casebox}

\paragraph{Well-Formed DAGs and Faster Convergence.}
RL also stabilizes DAG schema compliance, eliminating the retry storms caused by malformed transforms or unparseable Worker prompts. This is the qualitative root cause of the latency gap reported in \Cref{tab:reward_ablation}: ablating $R_{\text{fmt}}$ pushes latency from \textbf{381\,s} to \textbf{499\,s} due to repeated structural retries on rejected DAGs.

\begin{casebox}{Case G: Enterprise Survey Analysis (Format-Error Collapse)}
\textbf{Task.} Recommend a portfolio strategy from a 2023 enterprise survey and a 2024 quarterly report.

\smallskip
\textbf{After RL.} \emph{3 NARROWs, 224 facts, \textbf{0 Worker errors}.}

\smallskip
\textbf{Before RL.} \emph{8 NARROWs, \textbf{0 facts}, \textbf{459 cumulative Worker parse errors}.} The Driver's malformed queries cause Workers to raise parse exceptions on every shard:
\begin{quote}\footnotesize\ttfamily\raggedright
N1{=}63\ \ N2{=}72\ \ N3{=}49\ \ N4{=}65\ \ N5{=}36\ \ N6{=}83\ \ N7{=}56\ \ N8{=}35\ \ (errors)
\end{quote}

\textit{Observation.} The Before-RL Driver suffers complete planning collapse: every Worker invocation fails, no facts are extracted, and the orchestrator force-finalizes with empty context. The After-RL Driver's three well-formed NARROWs achieve full coverage with zero parse failures.
\end{casebox}

\begin{casebox}{Case H: Aggregate Convergence over 15 LongBench v2 (Long) Samples}
\centering
\small
\renewcommand{\arraystretch}{1.1}
\begin{tabular}{@{}lcc@{}}
\toprule
\textbf{Stages to final answer} & \textbf{After RL} & \textbf{Before RL} \\
\midrule
1 (immediate convergence)        & \textbf{15 / 15} & 7 / 15 \\
2--3                             & 0 / 15           & 2 / 15 \\
4--5 (hit max-stage cap)         & 0 / 15           & \textbf{6 / 15} \\
\bottomrule
\end{tabular}

\smallskip
\raggedright
\textit{Observation.} The Before-RL Driver hits the max-stage cap on 40\% of cases; the orchestrator must force a final answer via \texttt{<force\_final>yes</force\_final>}. The After-RL Driver converges in a single stage on every sample.
\end{casebox}

\end{document}